\documentclass{article} 
\usepackage{plain,times}
\usepackage{easyReview}
\usepackage{adjustbox}
\usepackage{amsmath}
\usepackage{amssymb}
\usepackage{booktabs}
\usepackage{bbold}
\usepackage{tikz} 
\usepackage{subcaption}
\usepackage{graphicx}
\usepackage{wrapfig}
\usepackage{tikz}
\usepackage{microtype}
\usepackage[inkscapelatex=false]{svg}
\usepackage{xurl}
\usetikzlibrary{positioning,arrows.meta,calc,decorations.pathreplacing, fit, backgrounds}

\usepackage{amsmath,amsfonts,bm}

\def\eqref#1{equation~\ref{#1}}

\def\1{\bm{1}}

\DeclareMathAlphabet{\mathsfit}{\encodingdefault}{\sfdefault}{m}{sl}
\SetMathAlphabet{\mathsfit}{bold}{\encodingdefault}{\sfdefault}{bx}{n}

\newcommand{\R}{\mathbb{R}}
\newcommand{\bx}{\mathbf{x}}

\newcommand{\bW}{\mathbf{W}}
\newcommand{\bR}{\mathbf{R}}
\newcommand{\bSa}{\mathbf{S}}   
\newcommand{\bB}{\mathbf{B}}
\newcommand{\bH}{\mathbf{H}}
\newcommand{\bM}{\mathbf{M}}
\newcommand{\bP}{\mathbf{P}}

\newcommand{\bDelta}{\boldsymbol{\Delta}}

\newcommand{\balpha}{\boldsymbol{\alpha}}
\newcommand{\sg}{\mathrm{sg}}

\newcommand{\softplus}{\sigma_+}

\usepackage{hyperref}

\hypersetup{
    colorlinks=true,
    linkcolor=blue,
    urlcolor=blue,
    citecolor=blue,
    pdftitle={Learning the Structure of Triangular Transport Maps},
    pdfpagemode=FullScreen,
    }

\title{Learning the Structure of Triangular Transport Maps}

\author{Morten Blørstad \thanks{Department of Informatics, University of Bergen.}
\And Pekka Parviainen \footnotemark[1] 
\And Berent Å. S. Lunde \footnotemark[1] \; \thanks{Equinor} 
}

\begin{document}

\maketitle

\begin{abstract}

Triangular transport maps provide a flexible approach to sampling-based probabilistic modeling, including density estimation, generative modeling, and Bayesian inference. They transform an unknown target distribution into a simpler reference through a monotone triangular map. The map structure is defined by a variable ordering and sparsity pattern, which together encode a directed acyclic graph. Map quality can depend strongly on this structure, yet finding a good structure is computationally expensive because each candidate generally requires fitting a different map. A central challenge is therefore to learn density and structure jointly, while keeping computation manageable as dimension grows.
We introduce \textit{Self-Structuring Transport Maps} (\textsc{SSTM}), which learn the map, ordering, and sparsity jointly. We use SoftSort to learn the variable ordering and $L_0$ gates to learn the sparsity, while preserving a triangular structure. To keep the map scalable, we use a monotone BatchEnsemble that shares one weight matrix across all map components through rank-one adapters.
Across synthetic and real data, jointly learning the structure and map gives better density estimates than estimating the structure first. When the structure is identifiable from the density, \textsc{SSTM} matches the density performance of a map fitted with the true structure and outperforms autoregressive flows. On large datasets, \textsc{SSTM} is competitive with autoregressive flows.

\end{abstract}

\section{Introduction}
\textbf{The problem.}
This work concerns the joint learning of a monotone triangular transport map and its structure from samples.
Triangular transport maps represent a target distribution by transforming it into a simple reference distribution \citep{marzouk2016sampling}, with applications in Bayesian inference \citep{marzouk2016sampling,baptista2024bayesian}, density estimation and generative modeling \citep{wang2022minimax,irons2022triangular}, experimental design \citep{huan2024optimal}, and data assimilation \citep{spantini2022}.
Their structure comprises a variable ordering, which determines a factorization into conditional densities, and a sparsity pattern, which restricts each conditional's dependence on preceding variables. An appropriate structure can yield parsimonious maps, improve sample efficiency, and expose dependencies \citep{ramgraber2025friendly}; a poor ordering can require a dense representation of the same distribution \citep{spantini2018}. However, a suitable structure is rarely known in advance. Searching over structures combines a combinatorial search over orderings with repeated map fitting: evaluating a candidate ordering can require fitting new, separately parameterized map components.
We ask whether the map and its structure can instead be learned together in a single optimization.

\textbf{Existing work.}
Most applications of triangular transport maps choose the structure before fitting the map. The ordering may be based on domain knowledge, selected from a small set of candidates, or constructed from spatial information \citep{ramgraber2023smoothing,bryutkin2025neural,schafer2021sparse}. Other methods adapt the structure during map estimation. \cite{baptista2024representation} fit each component with a sparse basis expansion, greedily adding terms under a fixed variable ordering. \cite{baptista2024sing} iteratively fit a transport map, estimates the Markov structure from the fitted density, and updates the ordering and sparsity pattern before refitting the map.

For density estimation, monotone triangular maps belong to the broader class of measure transport methods, including normalizing flows. Autoregressive flow layers and monotone triangular maps represent the same class of functions \citep{jaini2019,papamakarios2021review}. Monotonicity ensures invertibility, while the triangular structure gives efficient change-of-variable calculations through the Jacobian. Normalizing flows often compose many triangular layers, such as splines \citep{durkan2019nsf} or monotone neural networks \citep{huang2018naf}, with ordering permutations \citep{kingma2018glow} to increase expressiveness. Composition with ordering permutations loses the triangular structure and the direct conditional inference it enables. Extracting conditionals from a composed flow requires more general inference methods, such as MCMC in latent space \citep{cannella2021} or a separately trained conditional flow for each conditioning structure.

For structure discovery with triangular transport maps, \citet{xi2023triangular} show that a maximally sparse triangular map can identify the Markov equivalence class. Their approach searches over variable orderings and fits a map for each candidate, making the search increasingly expensive with dimension. They also identify reliable estimation of map sparsity as a practical challenge. \citet{izadi2024causal} avoid permutation search by recovering the ordering one variable at a time. At each step, they fit conditional models for the remaining
variables to identify the next root, then recover the graph in a separate step.

More general DAG-learning methods optimize graph structure directly. NOTEARS imposes a differentiable acyclicity constraint on a weighted adjacency matrix, with sparsity regularization and augmented-Lagrangian optimization \citep{zheng2018notears}. Its nonlinear extension represents each conditional with a separate nonlinear function \citep{zheng2020learning}. Graphical Normalizing Flows (GNF) extend this approach to density estimation by jointly learning a normalizing flow and a relaxed adjacency matrix that determines its conditioning structure \citep{wehenkel2021gnf}. GNF uses unconstrained monotonic neural networks (UMNNs) for monotonic
normalizers and the same sparsity and acyclicity machinery. This requires a sequence of optimization problems, which the authors report at least doubles training time.

Permutation-based methods have been used for causal discovery and avoid explicit acyclicity constraints by restricting edges to follow a learned ordering. \cite{cundy2021bcd} jointly learn an ordering and lower-triangular edge weights with variational inference under a linear-Gaussian structural equation model. \cite{charpentier2022dpdag} learn an ordering together with binary edges and nonlinear structural equations. They use Gumbel-Sinkhorn or Gumbel-Top-$k$ with SoftSort for the ordering, and a differentiable binary relaxation for the edges. \cite{kamkari2023ocdaf} learn a topological ordering from the likelihood of an affine autoregressive flow. Permutation-dependent masks share the flow parameters across orderings, while graph sparsity is recovered in a separate step.

\textbf{Our method.}

We introduce \textit{Self-Structuring Transport Maps} (\textsc{SSTM})\footnote{Code is available at: \url{https://github.com/MortenBlorstad/learning-the-structure-of-triangular-transport-maps}}
, which jointly learn the map, variable ordering, and sparsity from samples. The map components operate on subsets of the same feature space and share the goal of pushing target conditionals to the reference. We therefore treat them as a multi-task learning problem and parameterize them with a BatchEnsemble \citep{wen2020batchensemble}. The components share the main network parameters and have their own rank-one adapters. To ensure monotonicity in the diagonal variable, we introduce a monotone BatchEnsemble that restricts the diagonal variable to positive-weight paths while leaving the conditioning variables unrestricted. We furthermore learn the variable ordering with SoftSort \citep{prillo2020softsort} and use a hard permutation in the forward pass. Cumulative summation converts the permutation into nested triangular masks, which preserve the triangular structure throughout optimization. Stochastic $L_0$ gates \citep{louizos2018l0} learn sparsity within these masks. Straight-through estimators provide gradients to the ordering and gates, so the map and its structure are optimized jointly under the same transport objective.

\textbf{State-of-the-art joint estimation of structure and density.}
Few methods learn structure and density jointly as a factorization into conditionals. GNF is closest to our method and objective. \textsc{SSTM} matches or improves on GNF in density estimation, while requiring substantially fewer optimization steps. It also remains competitive with autoregressive flows while retaining a learned triangular structure. When the density favors a particular ordering, \textsc{SSTM} can recover much of the benefit of knowing the structure in advance.

\textbf{Practical advantages.}
\textsc{SSTM} reduces the cost of joint map and structure learning by sharing the main network across map components through a BatchEnsemble. The monotone parameterization preserves invertibility, while the learned ordering keeps the map triangular throughout optimization. Ordering, sparsity, and density are therefore learned jointly without an acyclicity constraint or a separate structure-learning stage.
\textsc{SSTM} also supports partial ordering constraints without fixing the full structure. For Bayesian inference, variables to condition on can be constrained to appear first, while the remaining ordering and sparsity are learned from data. This lets the user impose only the ordering constraints required by the inference task.

\section{Background}
\label{sec:background}
\subsection{Triangular Transport Maps}
\label{sec:bg:triangular}

Let $\bx \sim \pi$ be a sample from a $K$-dimensional target distribution and $\textbf{z} \sim \eta = \mathcal{N}(\mathbf{0}, \mathbf{I})$ follow a $K$-dimensional standard Gaussian reference. A triangular transport map $T: \R^K \to \R^K$ pushes $\pi$ to $\eta$ through the lower-triangular structure
\begin{equation}
T(\bx) = \begin{bmatrix} T_1(x_1) \\ T_2(x_1, x_2) \\ \vdots \\ T_K(x_1, \ldots, x_K) \end{bmatrix} = \begin{bmatrix} z_1 \\ z_2 \\ \vdots \\ z_K \end{bmatrix} = \textbf{z},
\label{eq:triangular_map}
\end{equation}
where each component $T_k: \R^k \to \R$ depends only on the first $k$ inputs and is strictly monotone increasing in its \emph{diagonal variable} $x_k$ \citep{marzouk2016sampling}. Monotonicity guarantees bijectivity, and the lower-triangular Jacobian yields an efficient log-determinant for change of measure computations:
\begin{equation}
\log \det \nabla_\bx T(\bx) = \sum_{k=1}^{K} \log \frac{\partial T_k}{\partial x_k}.
\label{eq:logdet}
\end{equation}

Given $N$ samples $\bx^{(i)} \sim \pi$, the map is fitted by minimizing the empirical KL divergence between $\pi$ and the pullback density $T^\sharp \eta$, which decomposes into $K$ independent per-component objectives \citep{baptista2024representation}:
\begin{equation}
\mathcal{J}(T) = \sum_{k=1}^{K} \mathcal{J}_k(T_k), \qquad
\mathcal{J}_k(T_k) = \frac{1}{N} \sum_{i=1}^{N} \left[
    \underbrace{\tfrac{1}{2} T_k(\bx^{(i)})^2}_{\text{mode-seeking}}
    \;-\; \underbrace{\log \frac{\partial T_k(\bx^{(i)})}{\partial x_k}}_{\text{spread-encouraging}}
\right].
\label{eq:transport_loss}
\end{equation}
The first term drives pushforward samples toward the Gaussian mode; the second term prevents collapse and encourages spread. This decomposition means each component can, in principle, be fitted independently.

\subsection{Why Ordering and Sparsity Matter}
\label{sec:bg:ordering}

When $x_j$ is conditionally independent of $x_k$ given $x_{1:k-1 \setminus j}$, the argument $x_j$ may be dropped from $T_k$, yielding a \emph{sparse} map. Sparsity reduces parameter count, improves sample efficiency, and makes the map more interpretable. The sparsity pattern depends on the variable ordering: different orderings correspond to different factorizations of $\pi$ into a product of conditionals, and some factorizations are much sparser than others \citep{spantini2018}.

\subsection{BatchEnsemble}
\label{sec:bg:be}

BatchEnsemble \citep{wen2020batchensemble} is a parameter-efficient approach to multi-task learning. It shares a single weight matrix $\mathbf{W} \in \R^{q \times p}$ across $K$ tasks, individualizing each through lightweight rank-1 adapters: an input adapter $\mathbf{r}_k \in \R^p$, an output adapter $\mathbf{s}_k \in \R^q$, and a bias $\mathbf{b}_k \in \R^q$. For nonlinearity $\phi$, task $k$ computes
\begin{equation}
\mathbf{h}_{\ell+1}^{(k)}
=
\phi\!\Big(
\mathbf{W}(\mathbf{h}_\ell^{(k)}\odot\mathbf{r}_k)
\odot\mathbf{s}_k+\mathbf{b}_k\Big).
\label{eq:be_single}
\end{equation}
Stacking adapters into matrices $\mathbf{R} \in \R^{K \times p}$, $\mathbf{S} \in \R^{K \times q}$, and $\mathbf{B} \in \R^{K \times q}$, the vectorized forward pass evaluates all $K$ tasks simultaneously:
\begin{equation}
\mathbf{H}_{\ell+1}
=
\phi\!\Big(
(\mathbf{H}_\ell\odot\mathbf{R})\mathbf{W}^\top
\odot\mathbf{S}+\mathbf{B}
\Big),
\label{eq:be_vectorized}
\end{equation}
at a total parameter cost of $pq + K(p{+}q)$ per layer, compared to $Kpq$ for $K$ independent networks.

\section{Self-Structuring Transport Maps}
\label{sec:method}

\textsc{SSTM} jointly learns a monotone triangular map, variable
ordering, and sparsity. Let $o=(o(1),\ldots,o(K))$ be a permutation of the $K$ variables, and index component $T_k$ by rank $k$ in this ordering. Its diagonal variable is $x_{o(k)}$, and it may depend on preceding variables. We encode these dependencies by a sparse triangular mask $\widetilde{\bM}$. Figure~\ref{fig:overview} summarizes the construction.

\tikzset{
    scell/.style={draw=black!40, line width=0.25pt, minimum size=0.26cm, inner sep=0pt},
    sdiag/.style={scell, fill=orange!85, draw=orange!90!black},
    sfree/.style={scell, fill=blue!18},
    soff/.style={scell, fill=white},
}
\newcommand{\minirow}[3]{%
  \tikz[baseline=-0.55ex]{
    \node[#1] at (0,0) {};
    \node[#2] at (0.26,0) {};
    \node[#3] at (0.52,0) {};}%
}

\begin{figure}[h]
\centering
\begin{adjustbox}{width=0.9\linewidth}
\begin{tikzpicture}[
    every node/.style={font=\small},
    cell/.style={draw=black!40, line width=0.3pt, minimum size=0.42cm, inner sep=0pt},
    free/.style={cell, fill=blue!18},
    off/.style={cell, fill=white},
    diag/.style={cell, fill=orange!85, draw=orange!90!black},
    keep/.style={cell, fill=teal!35},
    drop/.style={cell, fill=white, draw=red!55, line width=0.5pt},
    ref/.style={cell, fill=green!30},
    box/.style={draw=black!50, rounded corners=2pt, inner sep=5pt, align=center, font=\scriptsize},
    flow/.style={-{Latex[length=2mm]}, line width=0.9pt},
    dual/.style={-{Latex[length=2.4mm]}, line width=1.1pt, draw=black!75},
    sub/.style={-{Latex[length=1.6mm]}, line width=0.5pt, draw=black!60},
    fb/.style={-{Latex[length=1.6mm]}, line width=0.5pt, dashed, draw=gray!70},
    zoom/.style={dashed, gray!55, line width=0.4pt},
    scell/.style={draw=black!40, line width=0.25pt, minimum size=0.26cm, inner sep=0pt},
    skeep/.style={scell, fill=teal!35},
    soff/.style={scell, fill=white},
    sdiag/.style={scell, fill=orange!85, draw=orange!90!black},
    sfree/.style={scell, fill=blue!18},
    sdrop/.style={scell, fill=white, draw=red!55, line width=0.4pt},
    unit/.style={circle, draw=black!50, line width=0.4pt, minimum size=0.5cm, inner sep=0pt, font=\tiny},
    udiag/.style={unit, fill=orange!85, draw=orange!90!black},
    umono/.style={unit, fill=orange!85, draw=orange!90!black},
    ufree/.style={unit, fill=blue!18},
    uoff/.style={unit, draw=gray!40, fill=white, text=gray!60},
    uout/.style={unit, fill=green!30},
    emono/.style={draw=orange!90!black, line width=0.7pt},
    efree/.style={draw=blue!55, line width=0.45pt},
]
\def\cs{0.42}
\def\csm{0.26}

\node[box, align=left] (params) at (12.0, 3.45) {
    \textbf{learnable parameters}\\[2pt]
    $\boldsymbol{\Phi}$: ordering scores\\
    $\mathbf{A}$: gate affinity\\
    $\mathbf{W}$: shared weight\\
    $\mathbf{R}, \mathbf{S}, \mathbf{B}$: adapters
};

\begin{scope}[yshift=-6mm]
\node[font=\tiny, gray!70] at (4.25, 4.75) {computing the input mask};

\begin{scope}[shift={(3.5, 3.95)}]
    \node[skeep] at (-\csm, \csm) {};  \node[soff] at (0, \csm) {};   \node[soff] at (\csm, \csm) {};
    \node[skeep] at (-\csm, 0) {};      \node[skeep] at (0, 0) {};     \node[soff] at (\csm, 0) {};
    \node[skeep] at (-\csm, -\csm) {};  \node[skeep] at (0, -\csm) {}; \node[skeep] at (\csm, -\csm) {};
\end{scope}
\node[font=\tiny,right] at (3, 4.45) {$M$ (ordering mask)};

\begin{scope}[shift={(3.5, 3.0)}]
    \node[skeep] at (-\csm, \csm) {};  \node[soff] at (0, \csm) {};   \node[soff] at (\csm, \csm) {};
    \node[skeep] at (-\csm, 0) {};      \node[skeep] at (0, 0) {};     \node[soff] at (\csm, 0) {};
    \node[sdrop] at (-\csm, -\csm) {};  \node[skeep] at (0, -\csm) {}; \node[skeep] at (\csm, -\csm) {};
\end{scope}
\node[font=\tiny, right] at (3, 2.45) {$g$ (sparsity mask)};

\begin{scope}[shift={(4.8, 3.475)}]
    \node[sdiag] at (-\csm, \csm) {};  \node[soff] at (0, \csm) {};   \node[soff] at (\csm, \csm) {};
    \node[sfree] at (-\csm, 0) {};      \node[sdiag] at (0, 0) {};     \node[soff] at (\csm, 0) {};
    \node[soff] at (-\csm, -\csm) {};   \node[sfree] at (0, -\csm) {}; \node[sdiag] at (\csm, -\csm) {};
\end{scope}
\node[font=\tiny] at (4.8, 4) {$\tilde M = g \odot M$};

\draw[sub] (3.9, 3.95) to[out=0, in=150] (4.4, 3.6);
\draw[sub] (3.9, 3.0) to[out=0, in=210] (4.4, 3.35);

\draw[dashed, rounded corners=4pt, gray!60] (2.85, 2.25) rectangle (5.65, 4.65);
\end{scope}

\draw[zoom] (4.1, 0.4) -- (2.85, 1.65);
\draw[zoom] (4.1, 0.4) -- (5.65, 1.65);

\begin{scope}[shift={(7.75, 3.33)}, scale=0.33, every node/.style={transform shape}]
\begin{scope}[shift={(0,0.0)}]
\node[udiag] (ax1) at (0,2.2) {};
\node[uoff] (ax2) at (0,1.1) {};
\node[uoff] (ax3) at (0,0.0) {};
\node[umono] (am1) at (1.6,2.4) {};
\node[umono] (am2) at (1.6,0.7) {};
\node[umono] (af1) at (1.6,1.55) {};
\node[umono] (af2) at (1.6,-0.15) {};
\node[umono] (ag1) at (3.2,2.4) {};
\node[umono] (ag2) at (3.2,0.7) {};
\node[umono] (aq1) at (3.2,1.55) {};
\node[umono] (aq2) at (3.2,-0.15) {};
\node[uout] (az) at (4.4,1.15) {};
\node[font=\small, scale=2.2] at (-0.8,1.15) {$T_1$};
\node[font=\small, scale=2.2] at (5,1.15) {$z_1$};
\foreach \h in {m1,m2,f1,f2} \draw[emono] (ax1) -- (a\h);
\foreach \a in {m1,m2,f1,f2} \foreach \b in {g1,g2,q1,q2} \draw[emono] (a\a) -- (a\b);
\foreach \b in {g1,g2,q1,q2} \draw[emono] (a\b) -- (az);
\end{scope}
\begin{scope}[shift={(0,-3.2)}]
\node[ufree] (bx1) at (0,2.2) {};
\node[udiag] (bx2) at (0,1.1) {};
\node[uoff] (bx3) at (0,0.0) {};
\node[umono] (bm1) at (1.6,2.4) {};
\node[umono] (bm2) at (1.6,0.7) {};
\node[ufree] (bf1) at (1.6,1.55) {};
\node[ufree] (bf2) at (1.6,-0.15) {};
\node[umono] (bg1) at (3.2,2.4) {};
\node[umono] (bg2) at (3.2,0.7) {};
\node[ufree] (bq1) at (3.2,1.55) {};
\node[ufree] (bq2) at (3.2,-0.15) {};
\node[uout] (bz) at (4.4,1.15) {};
\node[font=\small, scale=2.2] at (-0.8,1.15) {$T_2$};
\node[font=\small, scale=2.2] at (5,1.15) {$z_2$};
\foreach \h in {m1,m2,f1,f2} \draw[efree] (bx1) -- (b\h);
\foreach \m in {m1,m2} \draw[emono] (bx2) -- (b\m);
\foreach \m in {m1,m2} \foreach \g in {g1,g2} \draw[emono] (b\m) -- (b\g);
\foreach \f in {f1,f2} \foreach \h in {g1,g2,q1,q2} \draw[efree] (b\f) -- (b\h);
\foreach \g in {g1,g2} \draw[emono] (b\g) -- (bz);
\foreach \q in {q1,q2} \draw[efree] (b\q) -- (bz);
\end{scope}
\begin{scope}[shift={(0,-6.4)}]
\node[ufree] (cx1) at (0,2.2) {};
\node[ufree] (cx2) at (0,1.1) {};
\node[udiag] (cx3) at (0,0.0) {};
\node[umono] (cm1) at (1.6,2.4) {};
\node[umono] (cm2) at (1.6,0.7) {};
\node[ufree] (cf1) at (1.6,1.55) {};
\node[ufree] (cf2) at (1.6,-0.15) {};
\node[umono] (cg1) at (3.2,2.4) {};
\node[umono] (cg2) at (3.2,0.7) {};
\node[ufree] (cq1) at (3.2,1.55) {};
\node[ufree] (cq2) at (3.2,-0.15) {};
\node[uout] (cz) at (4.4,1.15) {};
\node[font=\small, scale=2.2] at (-0.8,1.15) {$T_3$};
\node[font=\small, scale=2.2] at (5,1.15) {$z_3$};
\foreach \h in {m1,m2,f1,f2} \draw[efree] (cx1) -- (c\h);
\foreach \h in {m1,m2,f1,f2} \draw[efree] (cx2) -- (c\h);
\foreach \m in {m1,m2} \draw[emono] (cx3) -- (c\m);
\foreach \m in {m1,m2} \foreach \g in {g1,g2} \draw[emono] (c\m) -- (c\g);
\foreach \f in {f1,f2} \foreach \h in {g1,g2,q1,q2} \draw[efree] (c\f) -- (c\h);
\foreach \g in {g1,g2} \draw[emono] (c\g) -- (cz);
\foreach \q in {q1,q2} \draw[efree] (c\q) -- (cz);
\end{scope}
\draw[dashed, rounded corners=4pt, gray!60] (-1.5, -7) rectangle (5.5, 3);
\end{scope}
 
\draw[zoom] (7.2, 0.3) -- (7.6, 1.0);
\draw[zoom] (9.6, 0.3) -- (9.2, 1.0);

\begin{scope}[shift={(0.7, 0)}]
    \node[free] at (-\cs, 0) {\scriptsize $x_1$};
    \node[free] at (0, 0) {\scriptsize $x_2$};
    \node[free] at (\cs, 0) {\scriptsize $x_3$};
\end{scope}
\node[font=\scriptsize, align=center] at (0.7, -1.05) {target\ $\mathbf{x}\sim\pi$\\$\mathbf{x} \in \mathbb{R}^K$};

\draw[flow] (1.4, 0) -- node[above, font=\scriptsize] {tile $K\times$} (2.3, 0);


\begin{scope}[shift={(3.0, 0)}]
    \foreach \i in {-1, 0, 1} {
        \foreach \j in {-1, 0, 1} {
            \node[free] at (\j*\cs, \i*\cs) {};
        }
    }
\end{scope}
\node[font=\scriptsize] at (3.0, -1.05) {$\mathbf{X} \in \mathbb{R}^{K \times K}$};
\draw[flow] (3.7, 0) -- node[above, font=\scriptsize] {$\odot \tilde M$} (4.5, 0);


\begin{scope}[shift={(5.2, 0)}]
    \node[diag] at (-\cs, \cs) {};  \node[off] at (0, \cs) {};   \node[off] at (\cs, \cs) {};
    \node[free] at (-\cs, 0) {};    \node[diag] at (0, 0) {};    \node[off] at (\cs, 0) {};
    \node[off] at (-\cs, -\cs) {};  \node[free] at (0, -\cs) {}; \node[diag] at (\cs, -\cs) {};
\end{scope}
\node[font=\scriptsize] at (5.2, -1.05) {$\tilde{\mathbf{X}}$};

\node[box, fill=gray!5, align=left, font=\scriptsize] (be) at (8.5, 0) {
    shared $\mathbf{W}$, own adapters:\\[2pt]
    $z_1 = T_1(\minirow{sdiag}{soff}{soff};\ \mathbf{W}, \mathbf{r}_1, \mathbf{s}_1, \mathbf{b}_1)$\\[1pt]
    $z_2 = T_2(\minirow{sfree}{sdiag}{soff};\ \mathbf{W}, \mathbf{r}_2, \mathbf{s}_2, \mathbf{b}_2)$\\[1pt]
    $z_3 = T_3(\minirow{soff}{sfree}{sdiag};\ \mathbf{W}, \mathbf{r}_3, \mathbf{s}_3, \mathbf{b}_3)$
};
\node[font=\scriptsize] at (8.5, -1.05) {$\mathbf{z} = T(\tilde{\mathbf{X}};\ \mathbf{W}, \mathbf{R}, \mathbf{S}, \mathbf{B})$};

\draw[flow] (6.0, 0) -- (be.west);
\draw[flow] (be.east) -- (11.35, 0);

\begin{scope}[shift={(12.0, 0)}]
    \node[ref] at (-\cs, 0) {\scriptsize $z_1$};
    \node[ref] at (0, 0) {\scriptsize $z_2$};
    \node[ref] at (\cs, 0) {\scriptsize $z_3$};
\end{scope}
\node[font=\scriptsize, align=center] at (12.0, -1.05) {reference\ $\mathbf{z}\sim\eta$\\$\mathbf{z} \in \mathbb{R}^K$};

\node[box, fill=teal!8] (loss) at (12.0, 1.5) {
    $\mathcal{L} = \mathcal{J}(T) + \mathcal{L}_0$
};

\draw[flow] (12.0, 0.35) -- (loss.south);

\node[font=\scriptsize] at (6.6, 5.25) {pushforward\ \ $T(\mathbf{x}) = \mathbf{z}$};
\draw[dual] (1.0, 4.9) -- node[below, font=\scriptsize] {structure learning} (5.0, 4.9);
\draw[dual] (5.4, 4.9) -- node[below, font=\scriptsize] {density estimation} (12.2, 4.9);

\draw[dual] (12.2, -1.55) -- node[below, font=\scriptsize] {pullback\ \ $T^{-1}(\mathbf{z})=\mathbf{x}$\ \ (sequential bisection)} (1.0, -1.55);

\draw[fb] (loss.north) -- node[right, font=\scriptsize, gray!70] {$\nabla \mathcal{L}$} (params.south);

\end{tikzpicture}
\end{adjustbox}
\caption{\small Overview of \textsc{SSTM}. The forward map pushes
$\mathbf{x}\sim\pi$ to $\mathbf{z}\sim\eta$, while the inverse uses sequential bisection.  The learned ordering and sparsity gates form the sparse triangular mask $\tilde M = g\odot M$,  which is applied to the input.  Each of the $K$ masked rows is input to one map component $T_k$. The $K$ components share weights $\mathbf{W}$ and use rank-one adapters $(\mathbf{r}_k,\mathbf{s}_k,\mathbf{b}_k)$. The objective $\mathcal{L}=\mathcal{J}(T)+\mathcal{L}_0$ jointly optimizes the map, ordering, and sparsity. \emph{Cell colours:} orange, diagonal variable; blue, conditioning variable; white, masked off.}
\label{fig:overview}
\end{figure}
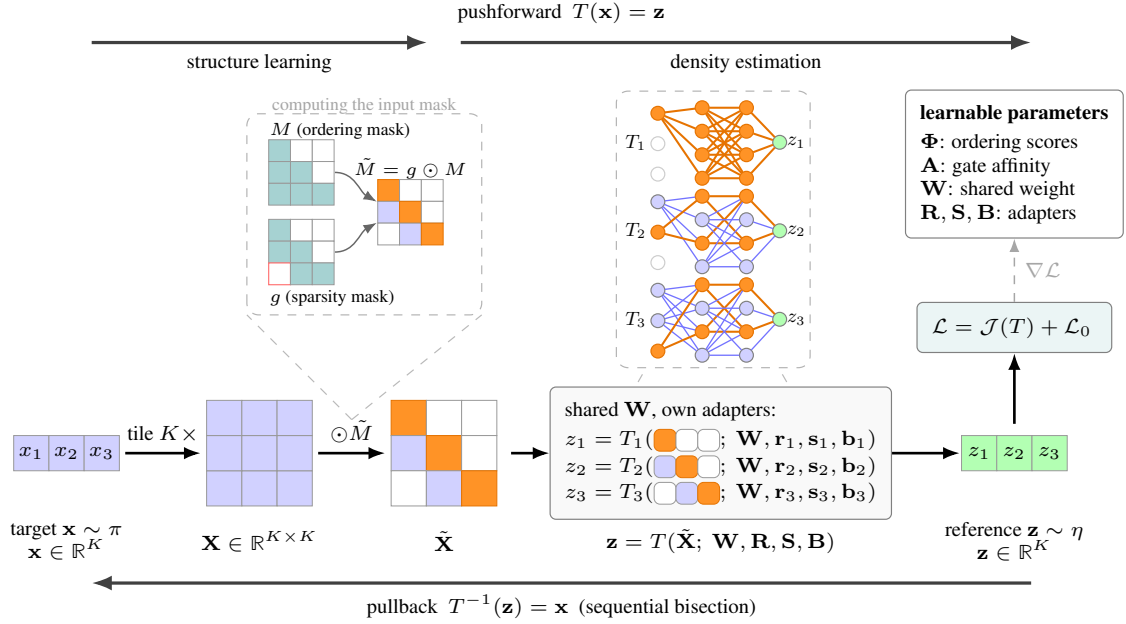

\subsection{Monotone BatchEnsemble}
\label{sec:mbe}
The $K$ map components solve the same transport problem on subsets of the same variable space. We therefore parameterize them with a BatchEnsemble that shares one weight matrix per layer and uses variable-indexed rank-one adapters routed according to the current ordering.

Each component must be monotone increasing in its diagonal variable. We enforce this by separating monotone and free paths through the network, illustrated in Figure~\ref{fig:overview} in orange and blue, respectively. The diagonal variable enters only the monotone path. Conditioning variables may also enter the free path. We use the monotonic activation of
\citet{runje2023constrained} in the hidden layers.

\textbf{Monotone BatchEnsemble layer.}
Let $V\in\{0,1\}^{K\times p}$ select monotone inputs and
$U\in\{0,1\}^{K\times q}$ select monotone output units. At the input
layer, row $k$ of $V$ selects the diagonal variable $x_{o(k)}$; at
hidden layers, it selects the monotone units of the preceding layer.
Let $\bW_+=\softplus(\bW)$. We split the input and output adapters as
\begin{align}
\bR_{\mathrm{mono}} &= \softplus(\bR)\odot V, &
\bR_{\mathrm{free}} &= \bR\odot(\mathbf{1}-V), &
\bSa_{\mathrm{mono}} &= \softplus(\bSa)\odot U.
\end{align}

For layer input $\bH_\ell$, the layer output is
\begin{equation}
\bH_{\ell+1}
=
\phi\!\left(
(\bH_\ell\odot\bR_{\mathrm{mono}})\bW_+^\top
    \odot\bSa_{\mathrm{mono}}
+
(\bH_\ell\odot\bR_{\mathrm{free}})\bW^\top
    \odot\bSa
+
\bB
\right).
\label{eq:monotone_be}
\end{equation}
Thus, the diagonal variable affects the output only through positive paths, while conditioning variables remain unrestricted.

\textbf{Memory-efficient weight normalization.}
We add weight normalization to the monotone BatchEnsemble
layer. Standard BatchEnsemble, member $k$ has effective weight matrix $\bW^{(k)}=\bW\odot(\mathbf{s}_k\mathbf{r}_k^\top)$. We normalize each effective row before applying the output adapter, so the shared weights and input adapters determine its direction and the output adapter determines its scale. Rather than forming the full $K\times q\times p$ tensor, we compute the row norms directly from the BatchEnsemble factors.

For monotone output units, the norm includes monotone and free input contributions; for free units, only the free contribution:
\begin{align}
\lVert\mathbf{w}_{\mathrm{mono}}\rVert^2
&= \bigl(\bW_+^2\, [\softplus(\bR)^2 \odot V]^\top
+ \bW^2\, [\bR^2 \odot (\mathbf{1}{-}V)]^\top\bigr)^\top, \\
\lVert\mathbf{w}_{\mathrm{free}}\rVert^2
&= \bigl(\bW^2\, [\bR^2 \odot (\mathbf{1}{-}V)]^\top\bigr)^\top, \\
\lVert\mathbf{w}\rVert
&= \sqrt{U \odot \lVert\mathbf{w}_{\mathrm{mono}}\rVert^2
+ (\mathbf{1}{-}U) \odot \lVert\mathbf{w}_{\mathrm{free}}\rVert^2
+ \epsilon},
\end{align}
where $\epsilon>0$ ensures numerical stability. The output adapters remain outside the normalization and act as component-specific gains. The normalized layer is
\begin{equation}
\bH_{\ell+1}
=
\phi\!\left(
\frac{
(\bH_\ell\odot\bR_{\mathrm{mono}})\bW_+^\top
    \odot\bSa_{\mathrm{mono}}
+
(\bH_\ell\odot\bR_{\mathrm{free}})\bW^\top
    \odot\bSa
}{
\lVert\mathbf{w}\rVert
}
+
\bB
\right).
\label{eq:monotone_be_norm}
\end{equation}

This retains $pq+K(p+q)$ non-bias weights per layer, compared with $Kpq$ for independent networks. All components and diagonal derivatives are evaluated in one vectorized forward pass by propagating derivatives alongside activations, avoiding $K$ separate backward passes.

\subsection{Learning the Variable Ordering}
\label{sec:ordering}

We parameterize the ordering by a score vector $\boldsymbol{\Phi}\in\mathbb{R}^K$. Sorting the scores in descending order gives the ordering $o$ and its permutation matrix $\widehat{\bP}\in\{0,1\}^{K\times K}$, where $\widehat{P}_{kj}=1$ if variable $j$ has rank $k$. SoftSort \citep{prillo2020softsort} gives a differentiable relaxation $\bP$. During training, we add Gumbel noise and sample multiple
orderings per input.

We use a straight-through estimator (STE) that combines the hard permutation in the forward pass with the SoftSort relaxation in the backward pass,
\begin{equation}
\bP^{\mathrm{STE}}
=
\widehat{\bP}
+
\bP-\sg(\bP),
\end{equation}
where $\sg(\cdot)$ denotes stop-gradient. We convert the permutation
into a nested triangular mask by cumulative summation,
\begin{equation}
\bM_{k,:}
=
\sum_{r=1}^{k}\bP^{\mathrm{STE}}_{r,:}.
\label{eq:ordering_mask}
\end{equation}
Row $k$ therefore selects the first $k$ variables in the current
ordering, and row $k$ of $\bP^{\mathrm{STE}}$ identifies the diagonal
variable of $T_k$. The forward pass always uses a hard triangular mask,
while gradients from the transport objective update
$\boldsymbol{\Phi}$ through SoftSort. Thus, the ordering is learned
without relaxing the triangular structure of the map. The newly introduced variable at rank $k$ is identified by
\begin{equation}
\bDelta_k
=
\bM_{k,:}-\bM_{k-1,:},
\qquad
\bM_{0,:}=\mathbf{0}.
\label{eq:delta}
\end{equation}
Thus, $\bDelta_k$ selects the diagonal variable of component $T_k$.

\subsection{Learning the Sparsity Pattern}
\label{sec:sparsity}

The triangular mask $\bM$ allows component $T_k$ to depend on every variable that precedes its diagonal variable in the ordering. We learn which of these conditioning variables are needed using stochastic
binary gates.

Let $\mathbf{A}\in\R^{K\times K}$ contain learnable gate affinities in the original variable coordinates. Since the rows of $\bM$ are indexed by rank, we map these affinities into the current ordering as
\begin{equation}
    \log\balpha = \sg(\bP^{\mathrm{STE}})\mathbf{A} ,
\end{equation}
where $\log\alpha_{kj}$ parameterizes the gate for variable $j$ in component $T_k$. The stop-gradient prevents the sparsity gates from updating the ordering through this operation.

During training, we sample
\begin{equation}
g^{\mathrm{soft}}_{kj}
=
\sigma\!\left(
\frac{\log u_{kj}-\log(1-u_{kj})+\log\alpha_{kj}}{\beta}
\right),
\qquad
u_{kj}\sim\mathrm{Uniform}(0,1),
\end{equation}
and use
$
g^{\mathrm{hard}}_{kj}
=
\mathbf{1}\!\left[
g^{\mathrm{soft}}_{kj}>\tfrac{1}{2}
\right]
$
in the forward pass. Gradients to the gate parameters are obtained with
the straight-through estimator
$
g^{\mathrm{STE}}
=
g^{\mathrm{hard}}
+
g^{\mathrm{soft}}
-
\sg(g^{\mathrm{soft}}).
$

The diagonal entries identified by $\bDelta$ are always retained, so the gates only prune conditioning variables. The resulting sparse triangular mask is
\begin{equation}
    \widetilde{\bM}=g^{\mathrm{STE}}\odot\bM .
\end{equation}
We encourage sparsity by penalizing the expected number of retained conditioning variables,
\begin{equation}
\mathcal{L}_0
=
\lambda_{\mathrm{sp}}
\sum_{k,j}
\sigma(\log\alpha_{kj})
\left(M_{kj}-\Delta_{kj}\right),
\label{eq:l0}
\end{equation}
where $\bM-\bDelta$ restricts the penalty to non-diagonal entries
allowed by the triangular mask.

\subsection{Joint Objective}
\label{sec:objective}

Given $N$ samples $\bx^{(i)}\sim\pi$, the sparse mask
$\widetilde{\bM}$ determines the inputs to each map component. The
transport objective becomes
\begin{equation}
\mathcal{J}(T)
=
\frac{1}{N}
\sum_{i=1}^{N}
\sum_{k=1}^{K}
\left[
\frac{1}{2}
T_k\!\left(
\widetilde{\bM}_{k,:}\odot\bx^{(i)}
\right)^2
-
\log\!\left(
\nabla_{\bx}T_k\!\left(
\widetilde{\bM}_{k,:}\odot\bx^{(i)}
\right)
\cdot\bDelta_k
\right)
\right].
\label{eq:joint_transport}
\end{equation}
The first term fits the pushforward to the Gaussian reference, while
$\bDelta_k$ selects the derivative with respect to the diagonal
variable of component $T_k$.

We optimize the transport objective together with the sparsity penalty,
\begin{equation}
\mathcal{L}
=
\mathcal{J}(T)
+
\mathcal{L}_0.
\label{eq:joint_objective}
\end{equation}
We additionally apply an $\ell_1$ penalty to the output-adapter gains through the optimizer:
\begin{equation}
\mathcal{R}_1
=
\lambda_1
\left(
\left\|\softplus(\bSa)\odot U\right\|_1
+
\left\|\bSa\odot(\mathbf{1}-U)\right\|_1
\right).
\end{equation}
The penalty acts on the positive gain $\softplus(\bSa)$ for monotone units and on the unrestricted gain magnitude for free units.

The map parameters, ordering scores $\boldsymbol{\Phi}$, and gate
affinities $\mathbf{A}$ are updated jointly. Straight-through
estimators provide gradients through the discrete permutation and
sparsity gates, while the forward pass remains a sparse triangular map.

\section{Experiments}
\label{sec:experiments}

We evaluate \textsc{SSTM} on synthetic and real data, comparing joint structure learning with random, known, and separately estimated structures, as well as flow-based density estimators.

\subsection{Experimental Setup}

\textbf{Datasets.}
We evaluate on three groups of datasets with different levels of ground truth: synthetic data with known density and structure, real data without known structure, and real data with a reference graph. 

\begin{figure}[h]
    \centering
    \includegraphics[width=\linewidth]{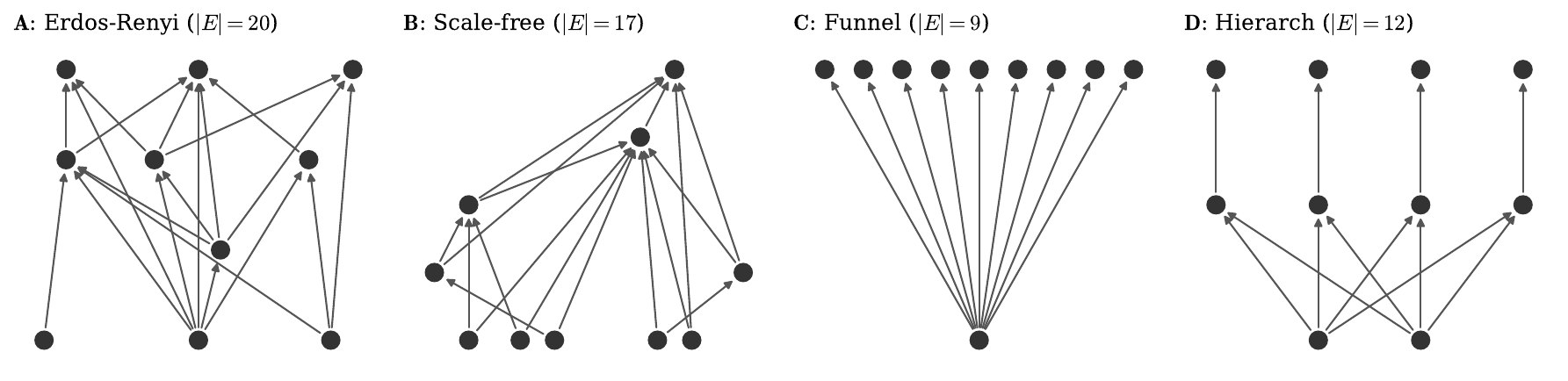}
    \caption{\small The four graph families. One generated DAG example per family at $K=10$ (edges oriented parent-to-child; $|E|$ = edge count). \textbf{A}:~Erd\H{o}s--R\'enyi, random edges. \textbf{B}:~scale-free, a few high-degree hubs. \textbf{C}:~Neal's funnel, a star with one root parenting all others. \textbf{D}:~two-level hierarchical, roots feed a middle layer that generates the leaves.}
    \label{fig:example_dags}
\end{figure}

For the synthetic experiments, we use six datasets with known directed acyclic graphs and tractable densities. Four are structural equation models generated as in \citet{zheng2020learning}, using linear and nonlinear mechanisms on Erd\H{o}s--Rényi \citep{erdos1960} and scale-free \citep{barabasi1999} graphs with $|E|=2K$ edges in expectation.  In these datasets, parents affect the conditional mean. The remaining two are Neal's funnel \citep{neal2003} and a two-level hierarchical model where root variables also control the scale of downstream variables. We vary $K\in\{10,20\}$ and $N\in\{200,1000\}$. \autoref{fig:example_dags} shows one example from each graph family. 

For real data without known structure, we use the four tabular UCI benchmarks of \citet{papamakarios2017maf,durkan2019nsf}: POWER, GAS, HEPMASS, and MINIBOONE, with their cleaning and splits. They range from $6$ to $43$ dimensions and from thirty thousand to $1.7$ million training points (\autoref{tab:density_hparams}).

For real data with a reference graph, we use the protein-signaling data of \citet{sachs2005}. It contains $11$ variables and a consensus graph with $20$ directed edges. We evaluate both the $853$ observational samples and the pooled $7{,}466$-sample dataset, which also includes interventional measurements. We use the $20$-edge reference graph throughout.

\textbf{Models.} We compare four triangular-map variants that share the same architecture and differ only in how their structure is obtained. Learned is \textsc{SSTM}, which learns the ordering and sparsity jointly with the map. Random fixes a random ordering and learns the sparsity. Two-stage first estimates the structure with nonlinear NOTEARS \citep{zheng2018notears,zheng2020learning}, using the authors' official implementation\footnote{\url{https://github.com/xunzheng/notears}}, and then fits the map with that structure fixed. Oracle fixes the true structure on synthetic data and the reference graph on Sachs. These variants compare joint structure learning with a random ordering, structure-first estimation, and known structure.

We also compare against flow-based models. Masked Autoregressive Flow (MAF) \citep{papamakarios2017maf} and Neural Spline Flow (NSF) \citep{durkan2019nsf} are density-only baselines, while Graphical Normalizing Flow (GNF) \citep{wehenkel2021gnf} jointly learns graph structure and density. We use the authors' official GNF implementation\footnote{\url{https://github.com/AWehenkel/Graphical-Normalizing-Flows}} on the synthetic and Sachs data. On the tabular benchmarks, we compare against the published GNF results.

\textbf{Setup.} All experiments use five seeds, with model selection based on validation data and final evaluation on held-out test data. We standardize using training-set statistics, removing marginal-variance cues. For the synthetic datasets, each seed generates a new graph and dataset with \(N\in\{200,1000\}\) training samples and \(5000\) validation and test samples. The same standardization is applied to the known true density when computing likelihoods. For the real data without known structure, we otherwise follow the preprocessing and data splits of \citet{papamakarios2017maf,durkan2019nsf}. For the Sachs data, each seed uses an \(80/10/10\) train, validation, and test split for both the observational and pooled data including interventions. Full architectures, optimization settings, and hyperparameters are given in Appendix \ref{app:training}.

\textbf{Metrics.} We evaluate density estimation by mean test negative log-likelihood (NLL). For synthetic datasets, where the true density is known, we report the gap: $
\mathrm{NLL}_{\mathrm{model}}-\mathrm{NLL}_{\mathrm{true}}.$

Where a true or reference graph is available, we evaluate structure using structural Hamming distance (SHD), precision, recall, and $F_1$ on directed edges. SHD counts missing, extra, and reversed edges. Precision is the fraction of predicted edges that are correctly oriented, recall is the fraction of true directed edges recovered, and $F_1$ is their harmonic mean.

\subsection{Experimental Results}

\begin{figure}[h]
    \centering
    \includegraphics[width=\linewidth]{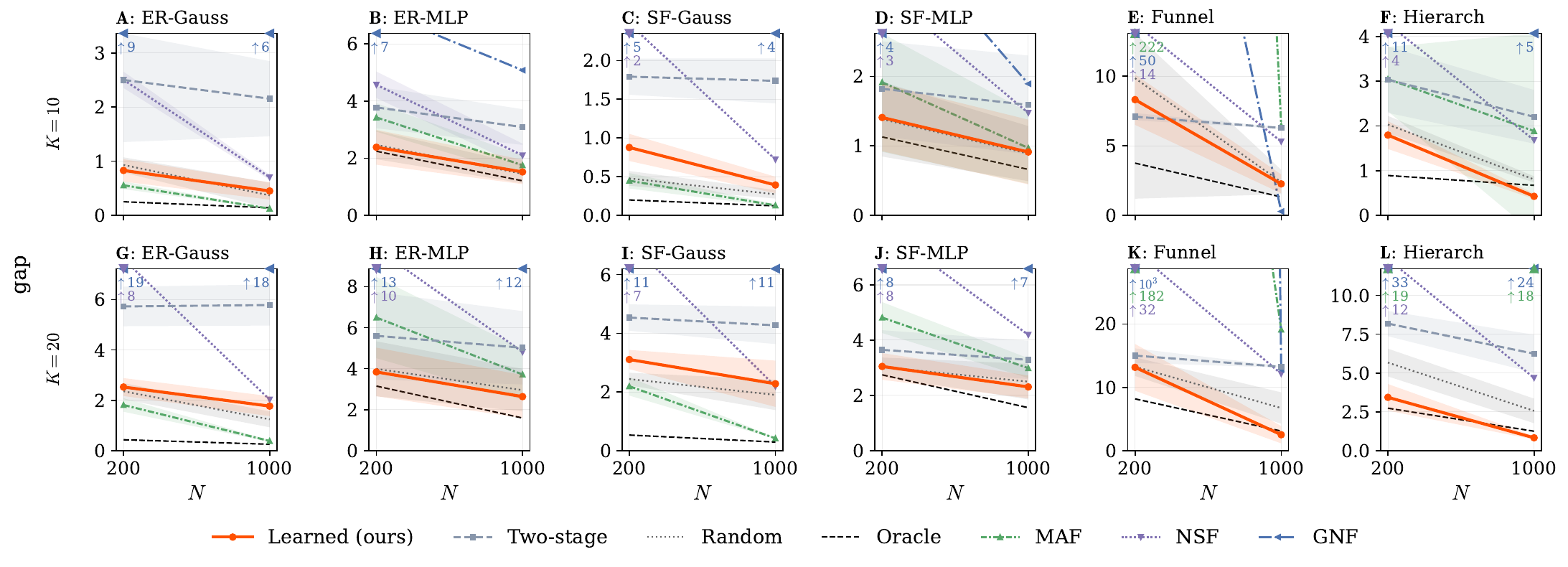}
    \caption{\small Gap to the true density per data-generating process (columns) and dimension $K$ (rows). Per-curve bands are $\pm 1$ standard deviation over five runs.}
    \label{fig:headline}
\end{figure}

\textbf{Known Synthetic DAGs}  Figure~\ref{fig:headline} compares density estimation across the 24 combinations of dataset, dimension, and training-set size. \textsc{SSTM} achieves a lower NLL gap than GNF in 23 of 24
combinations and than Two-stage in 23 of 24. The exceptions are both on the funnel at $K=10$: GNF performs better at $N=1000$
($0.28$ versus $2.26$), while Two-stage performs better at $N=200$ ($7.09$ versus $8.31$).

The benefit of learning the ordering depends strongly on the dataset. At $K=20$ and $N=1000$, the funnel gap is $6.73$ for Random, $2.52$ for \textsc{SSTM}, and $3.10$ for Oracle. On the hierarchical dataset, the corresponding gaps are $2.56$, $0.83$, and $1.25$. In contrast, Learned and Random are much closer on ER-Gauss, ER-MLP, SF-Gauss, and SF-MLP, although Oracle remains better. For example, on ER-Gauss at $K=20$ and $N=1000$, the gaps are $1.77$ for Learned, $1.24$ for Random, and $0.25$ for Oracle. At the same $K$ and $N$, GNF gives gaps of $12.81$ on the funnel, $23.96$ on the hierarchical dataset, and $18.19$ on ER-Gauss.

\textsc{SSTM} also outperforms MAF and NSF on the funnel and
hierarchical datasets for both dimensions and both training-set sizes. The differences can be large. On the funnel at $K=20$ and $N=200$, the NLL gaps are $13.16$ for \textsc{SSTM}, $31.54$ for NSF, $181.69$ for MAF, and $1850.85$ for GNF. MAF performs better than \textsc{SSTM} on the linear-Gaussian datasets.

Structure recovery shows a similar but not identical pattern. By
directed $F_1$, \textsc{SSTM} outperforms GNF in 22 of 24
combinations and Two-stage in 21 of 24. Full structure results,
including SHD, precision, and recall, are reported in the Appendix \ref{app:res}, Figure \ref{fig:radar_structure} and \ref{fig:radar_structure_n200}.

\textbf{Real data without reference graphs.} Table~\ref{tab:density_results} reports density estimation on the four tabular UCI benchmarks. \textsc{SSTM} achieves a lower NLL than MAF on three of four datasets, GNF on two, and NSF on one. NSF performs best on POWER and GAS, GNF on HEPMASS, and
\textsc{SSTM} on MINIBOONE.

\begin{table}[h]
  \centering
  \caption{\small Real-data density estimation on the UCI benchmarks, Mean NLL $\pm$ one standard deviation over five runs. Results followed by a star are copied from the \cite{wehenkel2021gnf} which reports over 3 runs.}
  \label{tab:density_results}
   {\scriptsize
  \begin{tabular}{@{}lrrrr@{}}
    \toprule
    & \textsc{Power} & \textsc{Gas} & \textsc{Hepmass} & \textsc{Miniboone} \\
    & $K=6$ & $K=8$ & $K=21$ & $K=43$ \\
    \midrule
    \textsc{SSTM} (ours) & $-0.527 \pm 0.020$ & $-10.854 \pm 0.039$ & $16.153 \pm 0.312$ & $9.925 \pm 0.137$ \\
    MAF                          & $-0.454 \pm 0.011$ & $-11.957 \pm 0.033$ & $17.058 \pm 0.250$ & $10.314 \pm 0.100$ \\
    NSF                       & $-0.650 \pm 0.003$ & $-13.031 \pm 0.023$ & $14.268 \pm 0.262$ & $10.300 \pm 0.089$ \\
    GNF$^\star$               & $-0.62 \pm 0.04$ & $-10.15 \pm 0.15$ & $14.17 \pm 0.13$ & $16.23 \pm 0.52$ \\
    \bottomrule
  \end{tabular}
  }
\end{table}

\begin{figure}[h]
    \centering
    \includegraphics[width=\linewidth]{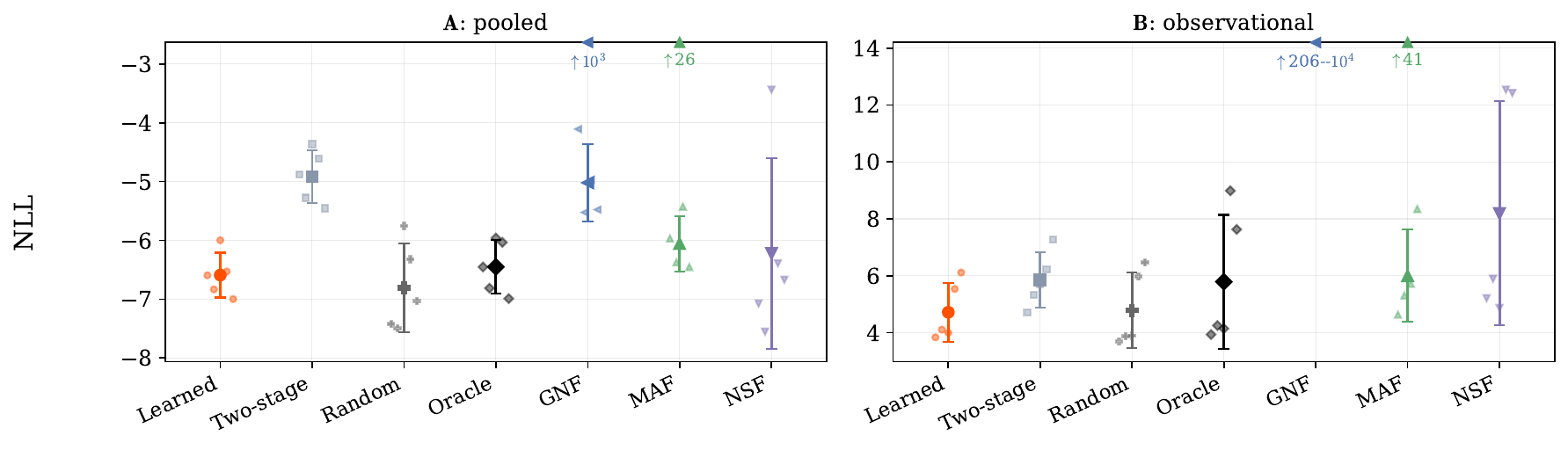}
    \caption{\small Density estimation on the Sachs data: test NLL on the pooled (\textbf{A}) and observational (\textbf{B}) subsets. 
    Small markers are single runs, bars the mean $\pm 1$ standard deviation over five runs. 
    Values beyond the axis range are marked at the frame with their value.}
    \label{fig:sachs_nll}
\end{figure}

\textbf{Real data with a reference graph.} 
Figure~\ref{fig:sachs_nll} reports density estimation on the Sachs
data. On the pooled data, Random has the lowest mean NLL, followed
closely by \textsc{SSTM} and Oracle. \textsc{SSTM} has a lower mean
NLL than MAF, NSF, GNF, and Two-stage. On the observational data,
\textsc{SSTM} has the lowest mean NLL, closely followed by Random,
while Oracle, Two-stage, and MAF have higher means. NSF has a still
higher mean and substantially larger variation. GNF diverges in all
observational runs and in one pooled run.

No learned structure method clearly outperforms Random. On the pooled
data, directed $F_1$ is similar for Random, GNF, and \textsc{SSTM}
($0.33$, $0.34$, and $0.29$). On the observational data, Random has
the highest directed $F_1$ at $0.38$, compared with $0.27$ for GNF
and $0.21$ for \textsc{SSTM}. Full structure results are reported in
the Appendix \ref{app:res}, Figure \ref{fig:sachs_radar}.

\textbf{Ablation of design choices.} We ablate the main design choices of \textsc{SSTM}. Figure~\ref{fig:ablation} in Appendix~\ref{app:ablation} shows that parameter sharing and weight normalization improve both density estimation and structure recovery. SoftSort performs similarly to Gumbel--Sinkhorn at substantially lower cost, while hard-concrete gates give similar density estimates but worse directed $F_1$.

\section{Discussion and Conclusion}
\label{sec:discussion}

The goal of this paper was to provide a triangular transport map that learns map, ordering and sparsity jointly from samples in a single optimization. Our experiments show that \textsc{SSTM} can jointly learn
a map and structure that provide accurate density estimates.

The experiments suggest that \textsc{SSTM} learns useful orderings
where they matter most, notably in hierarchical, funnel-type
distributions.
A suitable ordering simplifies the conditional maps, making them easier to learn from limited data; at fixed sample size,
this advantage may grow with dimension.
Elsewhere, alternative orderings perform similarly.
These findings suggest that \textsc{SSTM} favors good factorizations
whose complexity is supported by the available data.

Several design choices make the joint optimization practical.
\textsc{SSTM} preserves triangularity through the learned ordering and
monotonicity through the map parameterization. This avoids the
augmented-Lagrangian acyclicity optimization and UMNNs used by GNF.
In our experiments, GNF requires significantly more optimization steps.
Sampling also requires less computation per bisection step. Each step in \textsc{SSTM} requires one forward pass through the shared map, whereas GNF must numerically evaluate the UMNN integral using 20--40 function
evaluations. \textsc{SSTM} also reduce computational cost through parameter sharing across map components. The ablations show that this sharing, together with weight normalization, improves both density estimation and structure recovery. One possible explanation for the improvement is that the shared weights capture information common across map components, while the rank-one adapters handle component-specific changes as the structure changes. Parameter sharing and weight normalization may also provide useful regularization.

The main limitations concern optimization, scale, and evaluation.
Training requires tuning separate learning rates and schedules for the map, ordering, and sparsity parameters. Although the regularization strength adapts to the data through an EBIC-style schedule, more adaptive optimization could reduce tuning and make the method easier to use in practice. We also do not evaluate beyond $K=43$. Finally, although conditional sampling is an important motivation for triangular transport maps, we do not evaluate it directly. Future work should therefore focus on more adaptive training, scaling to higher dimensions, and evaluating the learned maps on conditional sampling tasks.  

In conclusion, \textsc{SSTM} shows that the map, ordering, and sparsity of a triangular transport map can be learned jointly from samples in a single optimization.

\clearpage

\bibliographystyle{plain}
\bibliography{references}

\clearpage  

\appendix
\section*{Appendix}

\section{Experimental Setup Details}
\label{app:training}

\subsection{Known Synthetic DAGs}

\textbf{\textsc{SSTM} and map variants.}
All four triangular-map variants use two hidden layers of width $32$ and train for $500$ epochs with batch size $64$. We use Adam with learning rate $10^{-2}$ for the map, ordering, and sparsity parameters, with an initial ramp-up followed by a constant learning rate and no decay. The sparsity gates use the straight-through estimator with penalty strength $\lambda_{\mathrm{sp}}=10^{-3}$. The ordering uses SoftSort with $m=10$ Monte Carlo samples per optimization step. The map is trained alone for the first $50$ epochs, after which ordering and sparsity learning begin. The SoftSort temperature is annealed from $0.3\sqrt{K}$ to $0.1$ over training. The $\ell_1$ strength $\lambda_1$ is an extended-BIC-style (EBIC) density prior \citep{chen2008ebic} that decays with the training-set size,
\begin{equation}
\lambda_1^\star = c\,\frac{\ln N}{2N}, \qquad c = 0.23 .
\end{equation}
The constant was set during development by fitting per-dataset schedules on validation diagnostics. The fitted schedules are EBIC-shaped with dataset-dependent constants between $0.03$ and $0.23$ using validation data. We use the largest, so a single schedule covers the most overfit-prone target.
For the models that learn sparsity (\emph{learned} and \emph{random}), $\lambda_1$ is ramped linearly from $0$ to $\lambda_1^\star$ over training, reaching its full value only once the SoftSort temperature has annealed to its floor (around epoch $390$ of $500$). This keeps $\lambda_1 \approx 0$ while the ordering and gates form, then applies the prior as the map refines. For the fixed-structure models (\emph{oracle} and \emph{two-stage}) there is no structure to protect, so $\lambda_1$ is held at $\lambda_1^\star$ throughout.

The \emph{two-stage} model estimates its structure with the MLP variant of NOTEARS, with their paper settings: hidden width $10$, penalties $\lambda_1 = \lambda_2 = 0.03$, edge threshold $0.5$, and at most $100$ iterations.

\textbf{Flow baselines.}
On the synthetic targets, NSF use the authors' implementation\footnote{\url{https://github.com/bayesiains/nflows}} with $5$ transforms, hidden width $128$, and two residual blocks per transform. MAF's affine scale is bounded to $(e^{-1}, e)$. NSF uses $4$ spline bins with tail bound $3$. Both train for $125$ epochs at batch size $64$ with Adam at learning rate $3 \times 10^{-4}$, dropout $0.2$, no weight decay, and a cosine schedule annealing the learning rate to zero. The configuration was chosen by validation search and is shared by both flows. We searched the number of transforms over $\{5, 10\}$, the hidden width over $\{32, 64, 128\}$, dropout over $\{0, 0.1, 0.2\}$, and weight decay over $\{0, 10^{-6}, 10^{-4}\}$. For MAF we additionally compared four parameterizations of its affine scale, from a contractive sigmoid to an unbounded softplus. The bounded scale in $(e^{-1}, e)$ was best and is the one reported.

\textbf{GNF.}
We use the authors' released implementation with the topology-learning architecture reported in their paper: a graphical conditioner with three hidden layers of width $100$, embedding size $30$, and an integrand network with three hidden layers of width $50$. We use batch size $100$ and Adam with learning rate $10^{-3}$. The main tuning concerned the augmented-Lagrangian schedule. The authors define the number of dual steps as the number of epochs between updates of the DAGness constraint. They use values between $10$ and $200$ across their experiments and note that increasing this value can improve performance at the cost of longer optimization. Their implementation default value is $100$ dual steps, but we found this required too many epochs to converge to a DAG. We therefore tuned the number of dual steps together with the training budget, using $10$ dual steps and $5{,}000$ epochs for $N=1000$, and $50$ dual steps and $20{,}000$ epochs for $N=200$. We use the augmented-Lagrangian progress factor $\gamma=1/4$ for $N=200$, and $\gamma=0.9$ for $N=1000$. Finally, we searched $\ell_1\in\{0,6,12\}$ and weight decay in $\{10^{-5},10^{-3},10^{-2}\}$ for a common setting across the synthetic targets, and use $\ell_1=12$ and weight decay $10^{-2}$.

\textbf{Reporting.}
All triangular-map models report the weights of the final epoch, with no checkpoint selection. The MAF, NSF and GNF report the epoch with the lowest validation negative log-likelihood, following their papers.

\subsection{Real Data without Known Structure}

\textbf{Budget and schedule.}
Every model trains at the per-dataset batch size and step budget of \citet[Table 5]{durkan2019nsf}, with the step budget converted to whole epochs. Validation, checkpointing, and the cosine schedule run per epoch rather than per $250$ steps as in their code. All models anneal the learning rate to zero with a cosine schedule. On the synthetic targets a paired comparison showed the cosine schedule does not change the results, so those experiments keep their original schedule.

\textbf{\textsc{SSTM}.}
We use the synthetic configuration above with three changes. The warmup is shortened from $50$ epochs to one, which still corresponds to more gradient updates because of the larger training sets. We use three hidden layers, with the width selected on validation data as shown in
Table \ref{tab:density_hparams}. We also searched the sparsity strength, but open gates performed best on validation data for all four datasets. We therefore keep the gates open and learn only the variable ordering. We also use a cosine schedule annealing the learning rate to zero.

\textbf{Reporting.}
\textsc{SSTM} reports the final epoch, which differs from its best-validation checkpoint by at most $0.03$ nats.

\begin{table}[h]
  \centering
  \caption{\small Hyperparameters of the density experiment on the UCI benchmarks, per dataset. Data: the standard splits of \citet{papamakarios2017maf}. Common: the per-dataset batch size and step budget of \citet[Table 5]{durkan2019nsf}, converted to whole epochs and shared by all models. Flows: their published per-dataset settings. \textsc{SSTM}: per-dataset size, chosen on validation.}
  \label{tab:density_hparams}
  \small
  \begin{tabular}{@{}lrrrr@{}}
    \toprule
    & \textsc{Power} & \textsc{Gas} & \textsc{Hepmass} & \textsc{Miniboone} \\
    \midrule
    \multicolumn{5}{@{}l}{\textit{Data}} \\
    Dimension $K$              & 6         & 8       & 21      & 43     \\
    Training points            & 1,659,917 & 852,174 & 315,123 & 29,556 \\
    Validation points          & 184,435   & 94,685  & 35,013  & 3,284  \\
    Test points                & 204,928   & 105,206 & 174,987 & 3,648  \\
    \midrule
    \multicolumn{5}{@{}l}{\textit{Common (all models)}} \\
    Batch size                 & 512       & 512     & 512     & 64     \\
    Step budget      & 400,000   & 400,000 & 400,000 & 250,000 \\
    Epochs                     & 124       & 241     & 651     & 543    \\
    \midrule
    \multicolumn{5}{@{}l}{\textit{Flows (MAF, NSF)}} \\
    Learning rate              & 0.0005    & 0.0005  & 0.0005  & 0.0003 \\
    Transforms                 & 10        & 10      & 10      & 10     \\
    Residual blocks            & 2         & 2       & 2       & 1      \\
    Hidden features            & 256       & 256     & 256     & 64     \\
    Spline bins (NSF)       & 8         & 8       & 8       & 4      \\
    Dropout                    & 0.0       & 0.1     & 0.2     & 0.2    \\
    \midrule
    \multicolumn{5}{@{}l}{\textit{\textsc{SSTM}}} \\
    Hidden features            & $256$ & $256$ & $256$ & $128$ \\
     \bottomrule
  \end{tabular}
\end{table}

\subsection{Real Data with a Reference Graph}

The triangular-map variants, MAF, and NSF use the synthetic configurations above. GNF runs at its authors' published Sachs configuration. The conditioner has three layers of width $100$ and embedding size $30$.
 The integrand network has three layers of width $50$. The $\ell_1$ strength is $12$. Some settings are not stated in their paper.
  We take them from the defaults of their released training script. These are Adam at learning rate $10^{-3}$, weight decay $10^{-5}$, 
  batch size $100$, and a constraint update every $100$ epochs.

 Some choices were needed. Their paper describes early stopping on the validation loss, but the released code does not implement it. 
 We use their code as is, using a fixed number of epochs instead. Their default is $10{,}000$ epochs. 
At this default, the smaller datasets did not converge to a DAG in any seed.
 We therefore scale the number of epochs so that each model trains for approximately the same number of gradient updates regardless of the dataset size.
  The pooled data trains for $10{,}000$ epochs, about $590{,}000$ updates. The observational subset trains for $85{,}000$ epochs, about $510{,}000$ updates.
 \textsc{SSTM} trains for $500$ epochs on the same splits, about $46{,}000$ updates.
 GNF reports the checkpoint with the best validation NLL among the states where its training has committed to an acyclic graph.

\newpage

\section{Additional Experimental Results}

\label{app:res}

\begin{figure}[h]
    \centering
    \includegraphics[width=\linewidth]{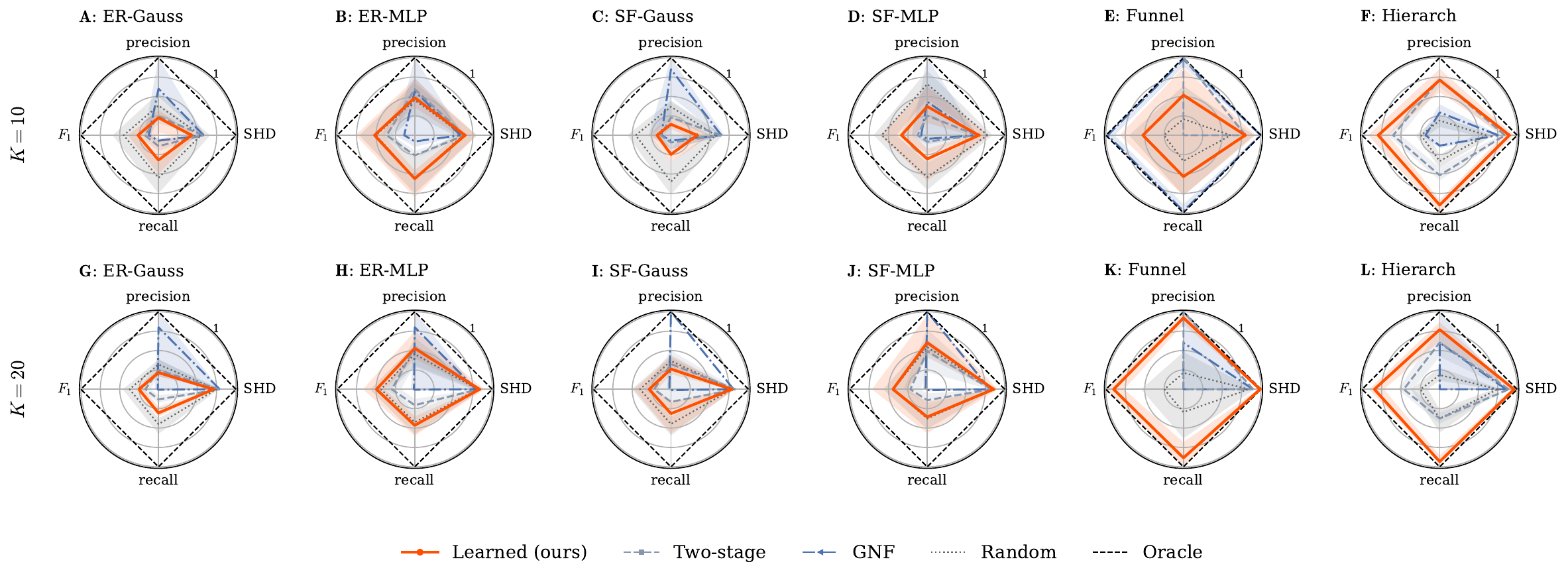}
    \caption{\small Structure recovery on the synthetic data at $N=1000$, per
    data-generating process (columns) and dimension $K$ (rows). Metrics are
     $[0,1]$ with $1$ best; SHD is inverted and normalized. Lines are means, bands the min to max over five runs.}
    \label{fig:radar_structure}
\end{figure}

\begin{figure}[h]
    \centering
    \includegraphics[width=\linewidth]{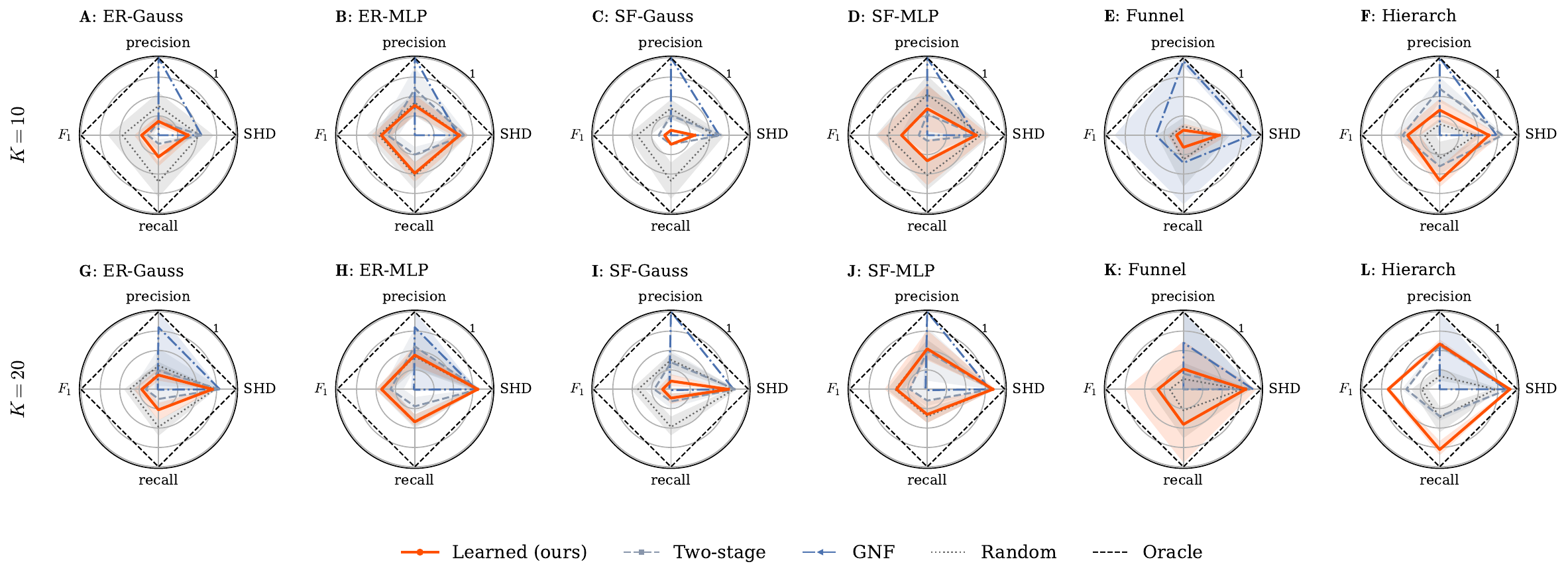}
    \caption{\small Structure recovery on the synthetic data at $N=200$, per
    data-generating process (columns) and dimension $K$ (rows). Metrics are $[0,1]$ with $1$ best; SHD is inverted and normalized. Lines are means, bands the min to max over five runs.}
    \label{fig:radar_structure_n200}
\end{figure}

\textbf{Synthetic structure recovery.}
Figures~\ref{fig:radar_structure} and~\ref{fig:radar_structure_n200} report the full structure metrics for the synthetic experiments. They complement the directed $F_1$ results in the main text with SHD, precision, and recall. Across metrics, the results show the same dataset-dependent pattern: \textsc{SSTM} recovers structure most reliably on the funnel and hierarchical targets, while recovery is weaker on the other graph families.

\begin{figure}[h]
    \centering
    \includegraphics[width=\linewidth]{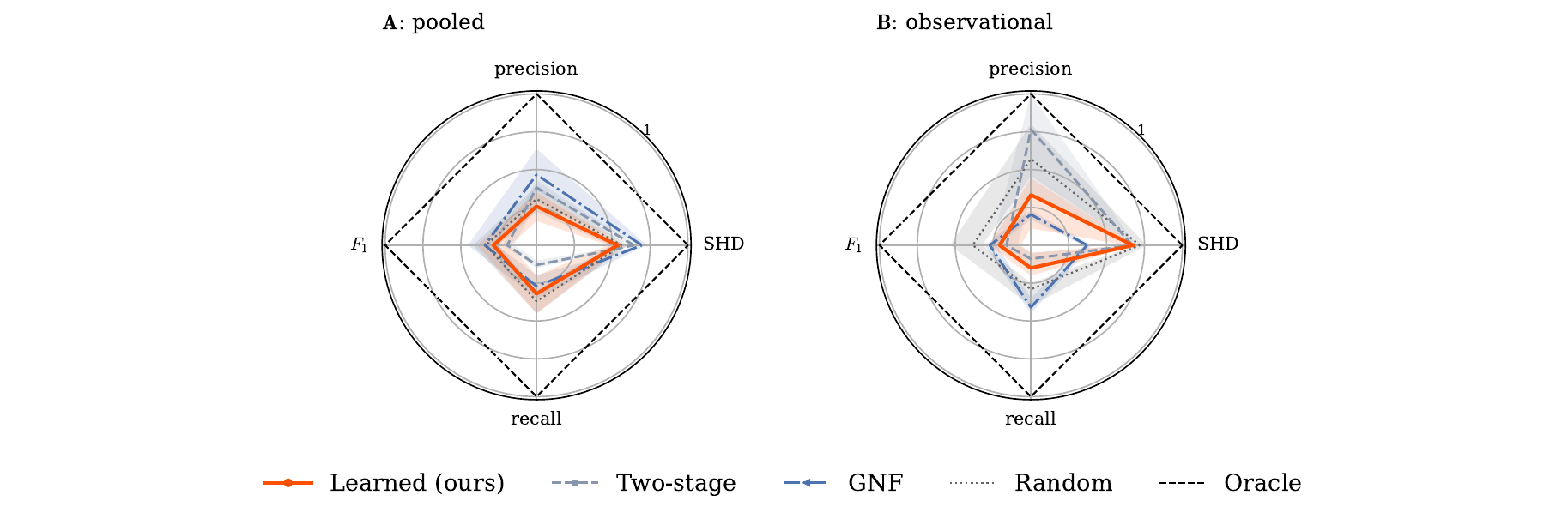}
    \caption{\small Structure recovery on the Sachs data against the 20-edge
    reference graph, on the pooled (\textbf{A}) and observational (\textbf{B})
    subsets. Metrics are scaled to $[0,1]$ with $1$ best; SHD is inverted and
    normalized. Lines are means, bands the min to max over five runs.}
    \label{fig:sachs_radar}
\end{figure}

\textbf{Sachs structure recovery.}
Figure~\ref{fig:sachs_radar} gives the full structure metrics against the Sachs reference graph. As discussed in the main text, no learned structure method consistently improves on random ordering with learned sparsity, despite their differences in density estimation.

\newpage

\subsection{Ablation of Design Choices}
\label{app:ablation}

\begin{figure}[h]
    \centering
    \includegraphics[width=\linewidth]{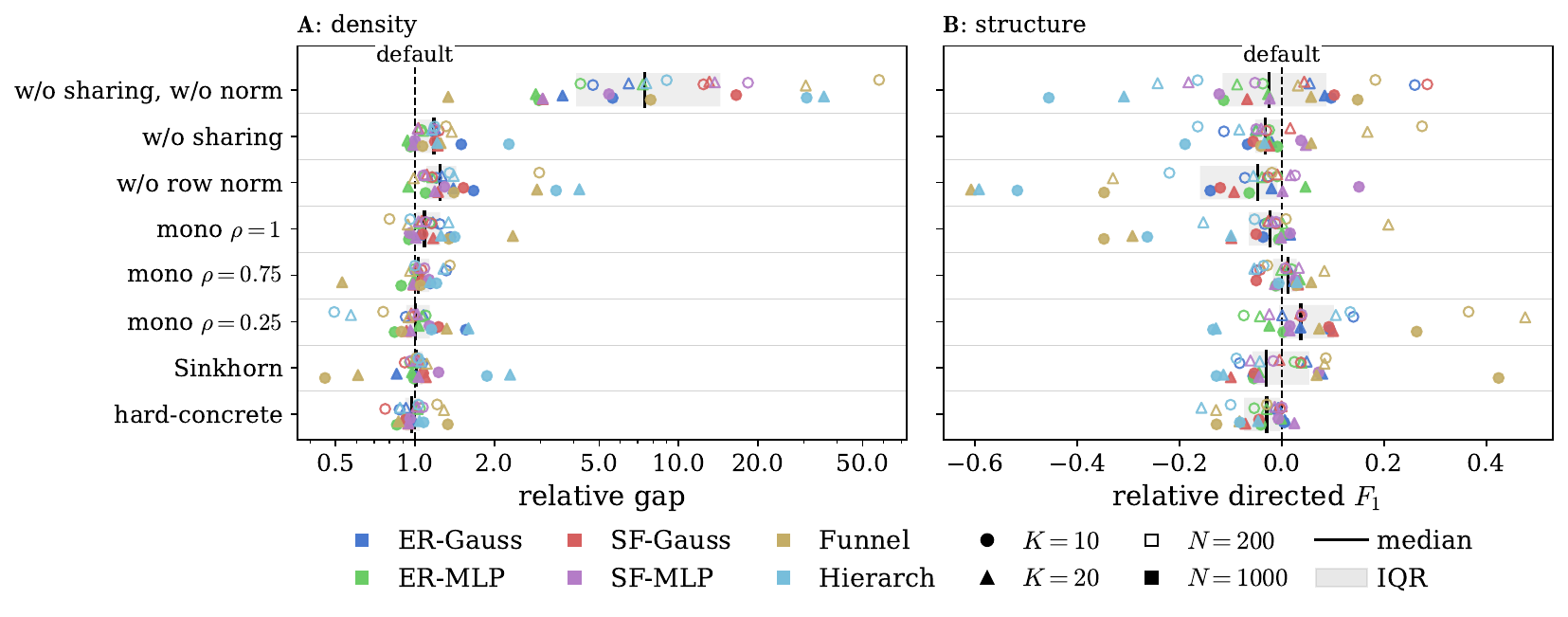}
    \caption{\small Method ablation: each variant changes one design choice of the default
    configuration and is compared against the default per dataset, dimension $K$, and
    training-set size $N$. Each marker is one such setting, averaged over five seeds
    (color = dataset, shape = $K$, fill = $N$). The black line is the median
    over settings and the grey band the interquartile range. The dashed line marks the
    default. \textbf{A}: gap to the true density relative to the default (ratio, log
    scale, $1$ = default, right = worse). \textbf{B}: directed $F_1$ relative to the
    default (difference, $0$ = default, right = better).}
    \label{fig:ablation}
\end{figure}

We ablate the design choices of \textsc{SSTM} one at a time. Each variant
changes one choice of the learned model and keeps everything else. We compare variant
and our default design on the gap and directed $F_1$ evaluated on test data.
Figure \ref{fig:ablation} shows the result.

\textbf{Parameter sharing and weight normalization.} Using BatchEnsemble reduces the number of parameters per layer from $Kpq \to pq + K(p{+}q)$. The parameter sharing reduces model expressivity.
Comparing the BatchEnsemble variant to fully independent map components, lets us measure whether parameter sharing affects density estimation and structure recovery.
Letting every component be its own model does not improve density estimation or structure recovery. It rather seems to hurt. The BatchEnsemble variant is better on density in 20 of 24 cells and on structure in 18 of 24 cells by directed $F_1$.
Sharing acts as an implicit regularizer across components.

weight normalization regularize and stabilizes training. Removing weight normalization makes training unstable and hurts both density estimation and structure recovery. 
It becomes even worse if we remove both parameter sharing and weight normalization.

\textbf{Monotonicity.} We make half of each layer's units monotone and leave the rest free. The free path is necessary. Letting all units be monotone hurts density estimation and structure recovery the most options. Letting 75\% of the units be monotone seem to equivalent to the default (50\% monotone). Letting 25\% of the units be monotone seems to improve structure recovery at the same density estimation performance.

\textbf{Ordering.} We relax the ordering with SoftSort because it is cheap and sufficient. SoftSort and Gumbel-Sinkhorn give similar performance.
 Gumbel-Sinkhorn does better on the funnel and worse on the hierarchical. But the big difference is computational cost. Gumbel-Sinkhorn`s is roughly 10x slower. SoftSort also only needs $K$ ordering parameters where Sinkhorn needs $K^2$.

\textbf{Sparsity.} We gate with a STE because it keeps the gates discrete,
whereas hard-concrete allows gate values between $0$ and $1$ both during
training and at inference. Hard-concrete gates give the same density but
worse structure recovery. They keep more edges in every setting (median
$29\%$ more) with the same recall, so the extra edges are false positives.

\section{Use of Large Language Models} We used a large language model as a coding assistant, to help debug code, and to set up experiments to run in parallel on a GPU server. We also used it to search for related work and to summarize existing papers, to draft and rewrite text based on the authors' own drafts, to edit for grammar and readability, and to improve figures, tables, and formatting. The authors designed and implemented the method, ran the analysis, and interpreted the results.  The authors verified all AI-assisted work and take full responsibility for the content of this paper.

\end{document}